\documentclass[fleqn,10pt]{wlscirep}
\usepackage[utf8]{inputenc}
\usepackage{listings}
\usepackage{booktabs}
\usepackage{multirow}
\usepackage{tabularx}
\usepackage{ltxtable}
\usepackage{array}
\usepackage{adjustbox}
\usepackage{geometry}
\usepackage{changepage}
\usepackage{graphicx}
\usepackage{lineno}
\usepackage{hyperref}

\usepackage{hyperref}
\usepackage[T1]{fontenc}

\title{The Collective Turing Test: Large Language Models Can Generate Realistic Multi-User Discussions}
\author[1]{Azza Bouleimen}
\author[2]{Giordano De Marzo}
\author[2]{Taehee Kim}
\author[1]{Nicolò Pagan}
\author[3]{Hannah Metzler}
\author[4]{Silvia Giordano}
\author[1]{Anikó Hannák}
\author[2,3,*]{David Garcia}
\affil[1]{University of Zurich, Department of Informatics, Zurich, 8050, Switzerland}
\affil[2]{University of Konstanz, Department of Politics and Public Administration, Konstanz, 78464, Germany}
\affil[3]{Complexity Science Hub, Vienna, 1030, Austria}
\affil[4]{University of Applied Sciences and Arts of Southern Switzerland, Department of Innovative Technologies, Viganello, 6962, Switzerland}

\affil[*]{david.garcia@uni-konstanz.de}


\begin{abstract}
Large Language Models (LLMs) offer new avenues to simulate online communities and social media.
Potential applications range from testing the design of content recommendation algorithms to estimating the effects of content policies and interventions.
However, the validity of using LLMs to simulate conversations between various users remains largely untested.
We evaluated whether LLMs can convincingly mimic human group conversations on social media.
We collected authentic human conversations from Reddit and generated artificial conversations on the same topic with two LLMs: Llama 3 70B and GPT-4o.
When presented side-by-side to study participants, LLM-generated conversations were mistaken for human-created content 39\% of the time. In particular, when evaluating conversations generated by Llama 3, participants correctly identified them as AI-generated only 56\% of the time, barely better than random chance. %In addition, when confronted to conversation generated by Llama 3, participants succeeded only 56\% of the time in identifying it as artificial, a bit over random guessing. 
Our study demonstrates that LLMs can generate social media conversations sufficiently realistic to deceive humans when reading them, highlighting both a promising potential for social simulation and a warning message about the potential misuse of LLMs to generate new inauthentic social media content.
\end{abstract}

\begin{document}
% \linenumbers 

\flushbottom
\maketitle
\thispagestyle{empty}

% \noindent Please note: Abbreviations should be introduced at the first mention in the main text – no abbreviations lists. Suggested structure of main text (not enforced) is provided below.

\section*{Introduction}

Agent-Based Modeling (ABM) approaches have been used to identify principles that drive the emergence of collective behavior such as political polarization~\cite{schweighofer2024raising}, segregation~\cite{schelling1971dynamic},
collective emotions~\cite{schweitzer2010agent}
cultural dynamics~\cite{axelrod1997dissemination}
and collective action~\cite{granovetter1978threshold}. 
While ABM can be used to illustrate generative sufficiency~\cite{epstein2012generative}, i.e. the minimal conditions for a social phenomenon to emerge, they tend to oversimplify human behavior ~\cite{flache2017models}, leading to limited application in policies or interventions. To increase the practical applicability of ABM in policy-making or system design, the dynamics of individual behavior and interactions between agents need to be grounded on empirical data ~\cite{windrum2007empirical}. However, data scarcity and the variance of human behavior complicates the task of model calibration and limits the ability of ABM simulations to generate predictions.

Recent advances in generative AI, in particular Large Language Models (LLMs), promise to narrow this gap as they demonstrate high role-play capabilities~\cite{grossmann2023ai,hamalainen2023evaluating,mei2023turing,bubeck2023sparks}. LLMs can enable a new kind of support for the study and simulation of collective behavior~\cite{larooij2025large, de2024ai} by complementing empirical data and allowing agents to communicate through real text or to decide agent actions based on LLM inference~\cite{xi2025rise}.

Social media are gaining significant traction as a prime application of social simulation. In light of the harm spreading on social media platforms~\cite{woolley2017computational,internet2020compaign} and their alarming real-world consequences~\cite{budak2019happened,guess2018selective,ferrara2017disinformation,yerlikaya2020social,islam2020covid,pierri2022online}, research on interventions and policies is needed to understand how to moderate the conversation and information flow on social media platforms. However, some intervention techniques~\cite{smith2025inoculation} fail to mitigate online social issues~\cite{chandrasekharan2022quarantined,trujillo2022make}, and can sometimes even cause further harm themselves~\cite{bail2018exposure,dias2020emphasizing,zollo2017debunking,horta2021platform}.

Given the ethical concerns of conducting large-scale experimentation on social media~\cite{kramer2014experimental}, generative ABM, i.e. the usage of generative agents in simulations, presents itself as an ideal safe and controlled environment where researchers and policymakers can design and test different intervention strategies. Simulating social media through generative ABM has become even more pressing in recent years in light of the restrictions to public data access that social media platforms have introduced~\cite{chan2024scaling}. 

Generative ABM has been previously used to simulate social media platforms and the interest they spark is on the rise. Early studies show that LLMs can link forming online social networks similar to those created by humans \cite{de2023emergence}. More recently, they have been employed to test how different feed algorithms affect the level of toxicity and cross-partisan interactions~\cite{tornberg2023simulating, gao2023s,ferraro2025agent}. Moreover, several different frameworks for simulating social networks have been developed~\cite{ferraro2025agent,yang2024oasis,rossetti2024social}. For example, recent work by Ferraro et al.~\cite{ferraro2025agent} provides a comprehensive framework for building specific personas as agents in a simulated social media environment. 
%The simulation comprises two cyclical components: a \textit{Reasoning Module} that emulates the agent's choice of action (generating original content, re-sharing content from other agents, remaining inactive) and an \textit{Interaction Module} that stores the agent's past behavior and manages the content to which the agent is exposed, i.e., the recommender system of the news feed. 
Despite this growing interest, there are still many open questions connected to the use of GABM. One of the most pressing is the need for novel validation approaches that show whether generative agents represent social media users, react in ways comparable to humans, and create overall group behavior that is akin to what humans do on social media. In other words, we still need to investigate the extent to which LLMs can mimic human behavior on social media. 

Some studies have given initial insights into how conversations can be simulated with LLMs. Park et al.~\cite{park2022social} introduced a prototyping technique to generate online forum discussions using GPT-3. For validation, they recruited crowdworkers to manually author comments on posts following authentic Reddit user comments. Both the crowdworker created conversations and the GPT-3 conversations were presented to a separate group of participants, who were asked to identify which conversation were AI-generated. The results showed that participants correctly identified AI-generated conversations 59\% of the time, indicating that GPT-3 can be suitable for simulating social media-like conversations. We note, however, that the crowdworker-created conversations used for comparison in this experiment were not authentic Reddit conversations. This limits the external validity of the experiment, as the control condition of crowdworker conversations is likely not representative of true online interactions. To improve on this approach, we suggest in our paper an experiment based on genuine Reddit conversations, created by active Reddit users.

Pagan et al.~\cite{pagancan} present a computational framework to systematically evaluate differences between LLM-generated social media and human replies. Their results reveal persistent divergences between human and LLM-generated replies, especially in affective and stylistic dimensions. However, they also showcase clear gains in realism from stylistic conditioning, context-aware prompting, and fine-tuning. While their approach is purely analytical, relying on BERT models and other Machine Learning (ML) systems to distinguish between LLM and human replies, our study employs an experimental design with human participants. This approach is complementary to~\cite{pagancan}, as insights provided by humans are likely to be different from what ML models detect. Besides the studies mentioned above, the overall reliability of generative ABM to simulate online conversations among various individuals remains under-explored. In a recent survey from Larooij and Törnberg~\cite{larooij2025large}, the authors state that the ``Validation of generative social simulations remain poorly addressed'' and that ``even the most rigorous validation fail to adequately evidence operational validity''. 

In this article, we present the results of a \textbf{Collective Turing Test}: Whether LLMs can generate the text of human-like conversations on social media. 75 years ago, Alan Turin described the Turing test to evaluate if a machine can generate a textual conversation that humans cannot distinguish from another human more than 30\% of the time \cite{turing1950computing}. Here, we extend that concept to a multi-user discussion between machines with a human as an observer, asking how often would LLMs generate conversations that humans cannot distinguish from other human-generated conversations.
We preregistered this study~\cite{preregistration} as the first experiment that uses real Reddit conversations to test the capability of LLMs to mimic human conversations on social media. Specifically, we present human participants with a set of Reddit posts followed by two side-by-side conversations of identical length and structure. One conversation is an authentic human conversation as it took place on Reddit, while the other is artificially generated by LLMs. Participants are asked to identify which is the human conversation. We generated AI conversations using two LLMs, GPT-4o and Llama 3 70B, and presented the participants with conversations of varying length. 
%At the end of the experiment, we asked participants to mention features that helped them in telling the AI conversation from the human one. 
Our first hypothesis is that the participants' success rate for conversations generated by Llama 3 70B will be lower than that of conversations generated with GPT-4o, i.e. that Llama 3 can generate more realistic conversations with the informality of Reddit users. Llama has been trained with internet-scale corpora that reportedly include high volumes of online forum-style dialogue, possibly aligning better with Reddit's tone and structure. Recently, it has been shown that Llama exhibits stylistic traits (such as tone, friendliness, and humor) that align well with human preferences in informal or social settings~\cite{dunlap2024vibecheck}. Second, we expect that longer conversations will be easier for participants to distinguish than shorter ones as longer threads provide more opportunities for participants to identify inconsistencies, unnatural flow, or semantic drift typical of AI-generated responses. However, this relationship may not necessarily be linear. 

%In the following, we show that Llama 3 70B showcased higher capabilities in appearing human to participants. We also demonstrate that even though the coefficients are not statistically significant, there exists a trend towards longer conversations being easier to spot than shorter ones.

\section*{Results}
% pre processing
A total of 251 participants took part in our study (see Methods for more details), generating 1741 annotations of pairs of authentic vs simulated discussions. We excluded all entries from 46 participants who failed our attention checks as well as the annotations corresponding to the attention checks. The remaining dataset contains 1205 annotations from 203 participants, satisfying the minimum sample size for 80\% statistical power. We define \textit{success rate} as the proportion of correctly identified human conversations by the participants. Overall, participants' success rate was 0.61. In other words, LLMs generated discussions that were mistaken to be human-made in 39\% of the cases.

\subsection*{The impact of model and conversation length on the success rate}
\begin{figure}
    \centering
    \includegraphics[width=\linewidth]{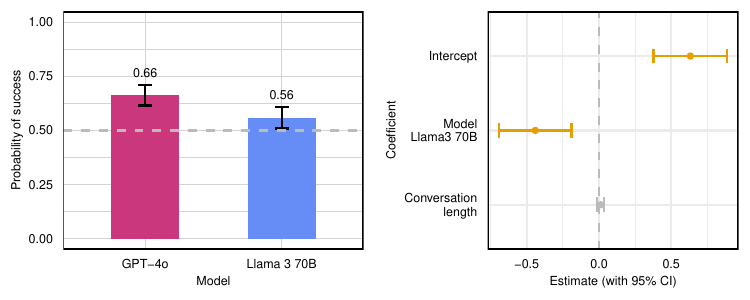}
    \caption{\textit{Left:} Predicted probabilities of correctly identifying the human conversation by model. \textit{Right:} Estimates of fixed effects in the logistic regression model with length as a numerical variable. 95\% confidence intervals are shown and bars in orange are significant at the 95\% level. %Intercept: $p-value = 1.14e-06$. Model Llama 3 70B: $p-value = 0.000562$. Conversation length: $p-value = 0.361143$.
    }
    \label{fig:base_model}
\end{figure}

In this section we report findings about our preregistered hypotheses. The results of the multilevel logistic regression are presented in Fig.~\ref{fig:base_model} and details about regression results are reported in SI Tab. S1.
There is a significant effect of the LLM used to generate AI conversations on the participants' ability to correctly identify human conversations. More precisely, participants had a lower success rate in identifying human conversations when they were presented with simulations generated with Llama 3 70B than with GPT-4o (0.56 vs 0.66,  $P < 0.001$). 
%The random intercept for users had a variance of 0.415 (SD = 0.644), indicating substantial differences in success rates between users. 
Based on these observations, we conclude that Llama 3 70B is better at mimicking the text of human social media-like conversations than GPT-4o. In fact, Llama 3, as a locally deployable model, may generate more organic and less polished responses than GPT-4o, which is optimized for reliability and safety in commercial applications as a chatbot. Furthermore, Llama 3’s training data likely includes social media conversations, given that it was developed by Meta. These characteristics may make Llama 3 outputs appear more human-like in informal online discussions.

\begin{figure}
    \centering
    \includegraphics[width=\linewidth]{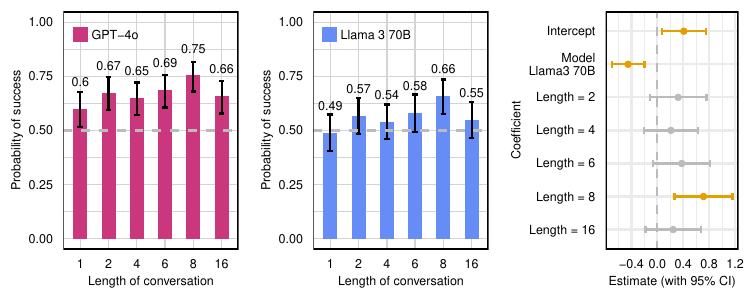}
    \caption{\textit{Left:} Predicted probabilities of correctly identifying the human conversation by model = GPT-4o and length as a categorical variable. \textit{Center:} Predicted probabilities of correctly identifying the human conversation by model = Llama 3 70B and length as a categorical variable. \textit{Right:} Plot of the fixed effect of the logistic regression when length is a categorical variable. Reference is Length = 1 comment. Coefficients with the orange CI are statistically significant.}
    \label{fig:factor_model}
\end{figure}

We also hypothesized that longer conversations are easier to identify than shorter ones. %Indeed, longer threads provide more opportunities for participants to identify inconsistencies, unnatural flow, or semantic drift typical of AI-generated responses. 
Results show no significant effect of conversation length on participant success rates ($p-value = \textit{0.36}$) when conversation length is included in the regression as a numerical variable, thus we cannot reject the null hypothesis that longer conversations are easier to spot than shorter ones. To further investigate the role of conversation length, we conducted an additional analysis treating length as a categorical variable to test for nonlinear effects. The results of this second model are presented on Fig.~\ref{fig:factor_model} and details can be found on Tab. S2 in the SI.

When treating conversation length as a categorical variable, we find that length generally has no significant effect on success rates, except for length = 8, which is associated with a significantly higher success probability compared to length = 1. Fig.~\ref{fig:factor_model} (left for GPT-4o conversations and right for Llama 3 70B conversations), shows the predicted probabilities of correctly identifying the human conversation by model and conversation length. For both models, results shows that as conversation length increases, success rates increase up to a certain point (in this case, length = 8), beyond which success rates decline (please note that only length = 8 is significant). This preliminary observation aligns with the intuition that very short conversations are harder to distinguish, while longer conversations provide participants with more opportunities to detect characteristic patterns of AI-generated content. Research shows that LLMs exhibit lower variability in their generated text compared to humans~\cite{zanotto2024human}. This trend holds as long as conversations remain moderately long. However, once conversation length reaches 16 posts, participant performance approaches random chance, suggesting that very long threads may make it harder to maintain focus. Given that only length = 8 coefficient is statistically significant, more investigations is needed in the direction of these observations.

\subsection*{Differences between human and AI conversations}
%analysis from post-survey justification of choice
As part of our preregistered exploratory analysis, we asked participants which cues helped them identify the AI-generated conversations after the annotation phase. Participants could freely provide their answers in a text box with few lines of space. Of the 203 participants who passed the attention test, 199 submitted a response to this question. To develop a codebook for classifying the responses, two researchers manually annotated an initial random subset of 50 answers. They identified three main categories: \textbf{Format}, \textbf{Style}, and \textbf{Content}. After discussion, the two researchers reached agreement on the scope of these categories and their respective subcategories. Details are provided in Tab.~\ref{tab:categories}. \textbf{Format} refers to visual cues in the text that can be noticed without reading it or knowing the conversational context. \textbf{Style} includes comments related to tone, voice, and stylistic choices. \textbf{Content} encompasses hints derived from the information itself, such as the level of explanation or argumentation provided.

\begin{table}[t]
\begin{tabularx}{\textwidth}{p{4cm}|p{6cm}|X}
\toprule
\textbf{Format}                                                                                 & \textbf{Style}                                                                                                                                                                             & \textbf{Content}                                                                                                                                  \\ \midrule
• \textit{emojis:} comments regarding the presence or absence of emojis in conversations.         & • \textit{tone:} comments noting the absence of sarcasm, irony, or humor in AI responses.                                                                                                  & • \textit{objectivity:} comments characterizing AI responses as purely factual without subjective elements.                                        \\
• \textit{structure:} comments addressing structural aspects of sentences and text organization. & • \textit{text formality:} comments characterizing AI conversations as unnatural, overly polished, professional, or robotic.                                                        & • \textit{explanations:} comments describing AI conversations as verbose or providing excessive detail and explanations.                           \\
• \textit{length:} comments concerning the relative length of conversational responses.           & • \textit{typos and slang:} comments observing that AI conversations contain fewer   typographical errors, grammatical mistakes, or domain-specific slang compared   to human conversations. & • \textit{controversy:} comments noting AI's tendency to avoid controversial or clear-cut positions and maintain neutral stances.                               \\
& • \textit{affective responses:} comments highlighting the absence of profanity or other forms of emotional language in AI conversations.                                                   & • \textit{conformity:} comments identifying lack of personality and excessive agreement with users in AI responses.                                \\
& • \textit{politeness:} comments describing AI text as excessively courteous or polite.                                                                                                     & • \textit{contextual misalignment:} comments indicating that AI replies were inappropriate for the given context.                                  \\
& • \textit{language diversity:} comments identifying repetitive text patterns, linguistic monotony, or insufficient response variation.                                                     & • \textit{general positivity:} comments observing that AI conversations exhibit greater positivity and enthusiasm compared to human conversations. \\
&                                                                                                                                                                                   & • \textit{authenticity:} comments identifying generic response patterns and absence of genuine personal opinions in AI conversations.     \\     \bottomrule   
\end{tabularx}
\caption{Categories and subcategories of the comments of participants on how they spotted the AI conversations.}
\label{tab:categories}
\end{table}

We restricted the qualitative analysis to the subset of participants with a success score of $\ge 0.8$ in identifying human conversations, resulting in 58 participants. Since these participants performed best in the task, their responses are more likely to convey reliable insights into the differences between AI and human conversations. Both researchers independently annotated this subset of answers according to the taxonomy described in Tab.~\ref{tab:categories}. A single comment could refer to multiple categories and subcategories. The Cohen's Kappa scores for inter-annotator agreement at both category and subcategory levels are reported in the Supplementary Tab. S4 of the SI online (Cohen's Kappa scores for the categories range between 0.66 and 0.75). After annotation, the two researchers met to discuss discrepancies and reached consensus on the final classification of the 58 comments. Table~\ref{tab:categories_subcategories} shows the percentage distribution of categories and subcategories.

% \usepackage{multirow}
\begin{table}
\centering
\begin{tabular}{l|c|l|c}
\toprule
\textbf{Category}        & \textbf{Category Percentage} & \textbf{Subcategory}    & \textbf{Subcategory Percentage} \\ \midrule
\multirow{3}{*}{Format}  & \multirow{3}{*}{17.5\%}      & emojis                  & 1.8\%                           \\
                         &                              & structure               & 7.0\%                           \\
                         &                              & length                  & 14.0\%                          \\ \midrule
\multirow{6}{*}{Style}   & \multirow{6}{*}{82.5\%}      & tone                    & 12.3\%                          \\
                         &                              & text formality          & 49.1\%                          \\
                         &                              & typos and slang         & 21.1\%                          \\
                         &                              & affective responses     & 12.3\%                          \\
                         &                              & politeness              & 14.0\%                          \\
                         &                              & language diversity      & 14.0\%                          \\ \midrule
\multirow{7}{*}{Content} & \multirow{7}{*}{49.1\%}      & objectivity             & 8.8\%                           \\
                         &                              & explanations            & 15.8\%                          \\
                         &                              & controversy             & 3.5\%                           \\
                         &                              & conformity              & 19.3\%                          \\
                         &                              & contextual misalignment & 1.8\%                           \\
                         &                              & general positivity      & 1.8\%                           \\
                         &                              & authenticity            & 12.3\%    \\ \bottomrule                     
\end{tabular}
\caption{Distribution of the categories and subcategories of hints mentioned in the participant's answers on what made them identify the AI-conversations.}
\label{tab:categories_subcategories}
\end{table}

Textual \textbf{style} features served as the primary indicator participants used when distinguishing human from AI conversations, with 82.5\% of responses mentioning at least one feature from this category. Within this category, most comments mentioned differences in \textit{tone} (49.1\% of total comments), \textit{text formality} (21.1\%), followed by \textit{politeness} and \textit{language diversity} (each at 14\%). This finding aligns with previous studies showing that language models, especially assistant-oriented ones, use considerably less toxic language than humans~\cite{pagancan}.\\
\textbf{Content}-related features ranked second, mentioned in 49.1\% of comments. The most frequent subcategories in this category were \textit{conformity} (19.3\%), \textit{explanations} (15.8\%), and \textit{authenticity} (12.3\%). This latter observation suggests that, in our experiment, the models did not showcase authenticity through personal stories and anecdotes, in comparison to real social media conversations. \\
\textbf{Format}-related indicators ranked third, appearing in 17.5\% of the participant comments. Most comments in this category mentioned response \textit{length} as an identifying feature, found in 14\% of all comments. Indeed, several participants stated things like ``... the answers were longer...'', ``Too verbose...'' and ``Large amount of text...'' about conversations they assumed to be AI-generated. This represents a particularly interesting observation given that LLMs were explicitly instructed to generate responses with approximately the same word count as their human counterparts. Considering that the second most frequent subcategory under \textbf{Content} was \textit{explanations}, we hypothesize that LLMs remain overly verbose compared to humans, even when producing comments of similar length on average. They may convey less information than humans within the same amount of text.

\section*{Discussion}
In this study, we conducted the first experiment investigating the capacity of large language models (LLMs) to mimic the collective behavior of humans on social media platforms using real Reddit data. To generate AI conversations, we used a minimal prompting strategy, asking the models to behave like Reddit users without assigning any specific personas (the exact prompt is provided in the Supplementary Information). Despite this simplicity, participants failed to identify the human conversation in 39\% of cases. This exceeds the 30\% misclassification threshold traditionally associated with passing the Turing Test, suggesting that even basic prompting can yield conversations that convincingly imitate human interactions.

Notably, conversations generated by \textbf{Llama 3 70B} were significantly harder to distinguish from human ones than those produced by \textbf{GPT-4o}, pointing to meaningful differences in how specific models simulate human-like discourse. These findings are \textit{promising} for researchers seeking to use LLMs in ABMs to simulate social systems more realistically, as they suggest that agents powered by LLMs can convincingly emulate human dialogue. However, the results are also \textit{concerning}, as they highlight the risk of LLM-generated content blending into genuine online discourse undetected. This could be exploited to create artificial consensus or engineer social polarization through coordinated ``AI swarms,'' as previously suggested in the literature~\cite{schroeder2025malicious}.

Surprisingly, \textbf{conversation length} did not significantly impact participants' ability to distinguish between AI and human threads. However, we observed an \textit{inverse-U relationship}, where detection accuracy initially improved with conversation length peaking at 8 comments before declining. This non-linear pattern suggests that longer conversations, up to a certain length, may introduce enough natural-sounding variance to mask the LLM origin, or may lead to participant fatigue. Both hypotheses warrant further investigation.

Our \textbf{qualitative analysis} of participant responses revealed several recurring cues used to assess authorship. Many noted that AI-generated conversations were overly \textit{polite}, \textit{formal}, and devoid of slang or profanity, features that are commonplace in authentic Reddit discourse. This suggests a limitation of current LLMs in simulating toxic or emotionally charged behavior, which is crucial for studying online conflict or misinformation. Participants also observed a lack of \textit{authenticity}, emotional nuance, and personal storytelling in AI responses, which reduced the realism of emotional or persuasive discourse. Additionally, a frequent comment was the tendency for AI users to \textit{over agree} with one another, which could compromise the validity of simulations focused on disagreement or polarization.

These findings highlight both the \textbf{potential and the limitations} of current LLMs in replicating human behavior in online settings. Careful prompt engineering, model selection, and possibly persona-driven prompting will be necessary to improve realism, especially in simulations of contentious or emotionally rich topics.

%Our study investigates the extent to which LLMs can mimic collective conversations. However, the limited scope of our experiment does not allow the observations to be generalized to all social media platforms. For instance, Instagram and TikTok feature markedly different styles of social interaction compared to Reddit or Twitter/X. Another limitation lies in the restricted range of models tested.

\section*{Conclusion}

Our findings offer three key takeaways:

\begin{enumerate}
    \item \textbf{LLMs can convincingly simulate human discourse on social media.} Across models and conversation lengths, human evaluators mistook LLM-generated Reddit conversations for human-authored ones in 39\% of cases, surpassing the Turing Test threshold. This highlights the potential of LLMs for generative agent-based modeling (ABM).
    
    \item \textbf{Model selection significantly affects performance.} Llama 3 70B outperformed GPT-4o in producing more human-like conversations. Based on these results, we recommend Llama 3 70B for use in simulations aiming to replicate naturalistic human interaction.
    
    \item \textbf{Limitations persist in simulating emotional and polarizing dynamics.} Although LLMs often pass as human, they struggle to replicate emotionally charged discourse, personal storytelling, or polarization, factors that are central to many social media phenomena.
\end{enumerate}

Future work should extend this ``Collective Turing Test'' approach to include additional LLMs, such as proprietary models like Claude and open-weight models like Mistral. These evaluations could serve as benchmarks for selecting models in social simulations. Moreover, the study can be replicated in other languages, leveraging Reddit's multilingual data. Another promising direction involves using multiple persona-driven LLM agents in conversations to assess whether this setup improves authenticity, fosters personal storytelling, or generates more emotionally intense or controversial discourse.

In summary, LLMs exhibit substantial capabilities in mimicking online human behavior, but important limitations remain. Researchers must apply these tools critically, especially when simulating sensitive phenomena such as polarization, misinformation, or emotional contagion. We hope this work supports the growing community of computational social scientists by offering empirical benchmarks, practical model comparisons, and methodological guidance for conducting LLM-based social simulations.

\section*{Methods}
\label{sec:methods}
To build and conduct our experiment, we first collected real human data from Reddit and then generated parallel AI conversations. After that, we built the experiment environment and ran the experiment with online participants, following our preregistration.

\subsection*{Collecting and generating data for stimuli}
We sampled Reddit data from Pushshift dataset, which contains an exhaustive compilation of public Reddit discussions up to December 2023~\cite{baumgartner2020pushshift}. We selected conversations from December 2023 to avoid including data potentially present in the training sets of GPT-4o or Llama 3 70B, as their training cutoffs are Oct. 1st 2023. and Dec. 1st 2023, respectively~\cite{OpenAI2025,llama}. From the Reddit posts we collected, we selected 16 posts spanning 8 different topics: food, leisure, opinions, relationships, technology, movies, health, and sports. The exhaustive list of selected Reddit posts is presented in Supplementary Tab. S3 online. For each post, we scraped the first 20 comments as they appear to a logged-in user. Reddit's comment ranking algorithm is user-independent and relies on comment-specific metrics such as upvotes and downvotes~\cite{Reddit}. Consequently, all logged-in users see the comments in the same order, allowing us to replicate what typical Reddit user experience. We anonymized the data by replacing usernames and profile pictures with fictional alternatives. From the scraped comments, we extracted their conversational structure: which user replied to which other user under which thread of comment, and word counts for each comment. In other words, we kept the tree-structure of the conversation fixed, while we replaced each human comment with an AI generated one. In doing so, we provided the LLMs with the length (in number of words) of the original comment to preserve the length of comments. This way, we only focus on the conversation content, while leaving structure and length fixed.

To generate the artificial conversations, we prompted both models with the text reported in Sec. Model Prompt of the SI. We did one round of refinement of the prompt following feedback from a pilot study and left the prompt unchanged after we preregistered our experiment. For this study, we did not attempt to create distinct personas when prompting the LLMs. All LLM instances shared the same standard model configurations with additional instruction on behaving as Reddit users. No conversational examples were previously provided to the models. For each comment to generate, we created a new LLM instance and provided it with the prompt, the post text, and all comments generated so far by previous LLM instances. Specifically, we wrote:
\begin{verbatim}
    Write a reply with about {word_count} words to comment ID {in_reply_to_id}.
\end{verbatim}

The model was then tasked with generating text of approximately the same length as the corresponding human comment (\texttt{word\_count}), either in reply to the corresponding user or as a direct comment on the post (represented by \texttt{in\_reply\_to\_id}).
We generated conversations using GPT-4o and Llama 3 70B in early Dec. 2024. For each model, we used three different temperatures: 0.2, 0.7, and 1.2. The same conversation could then be shown to participants at 6 different lengths, consisting of the first 1, 2, 4, 6, 8, or 16 comments from the generated conversations. Following this procedure we obtained 574 different stimuli combining posts with AI versus human conversations for presentation to participants (2 LLM models × 3 temperatures × 6 lengths × 16 posts). These conversations were generated once before our experiment preregistration and were not regenerated afterwards.

\subsection*{Experimental setting}
We used the same HTML layout as Reddit's webpage to generate images of both AI and human conversations. For the annotation task, we relied on Potato~\cite{Potato}, a Python library for building annotation systems, which we adapted to the needs of our experiment. Each participant first viewed an information page about the experiment (See Sec. Information page in SI). This page clearly stated that participants will see two conversations in every task, and that one is AI-generated while the other is human. After that, the experiment started with a pre-screening survey where participants provided consent and answered two basic attention-check questions (See  Supplementary Fig. S1). Participants who failed either question were not allowed to proceed and were directed to an exit page, while those who succeeded advanced to the annotation phase. In this phase, each participant completed 6 annotations and 1 additional attention test (An example is provided in Supplementary Fig. S2). To prevent redundancy, we extended the Potato library with a custom sampling algorithm ensuring that no participant saw the same Reddit post more than once, even if presented with different conversation lengths, LLM models, or temperature settings. 

\begin{figure}
    \centering
    \includegraphics[width=0.7\linewidth]{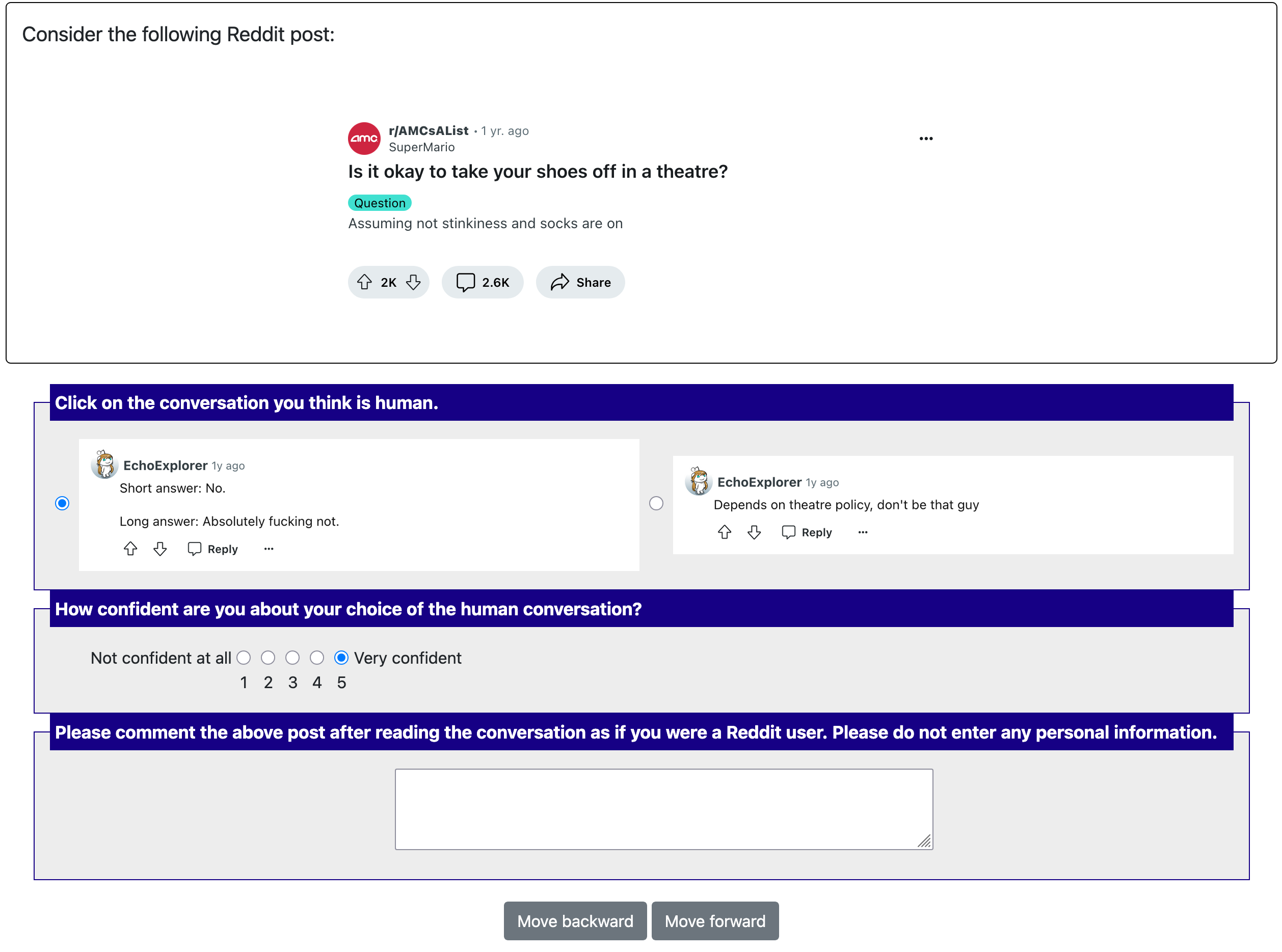} %screenshot_annotations.png
    \caption{Example of annotations shown to participants.}
    \label{fig:screenshot}
\end{figure}

To avoid acquiescence bias~\cite{billiet2000modeling}, every annotation task included two conversations, side-by-side, one human and one AI generated. An example of the interface presented to participants is shown in Fig.~\ref{fig:screenshot}. The two conversations have the same number of comments from the same number of users involved, having the same usernames and profile pictures. The only thing that differs between the human and the AI conversations is the text of the comments, which was made to be approximately the same number of words. We asked the participant to select which conversation was human-made. When doing, so the other conversation was greyed out. Participants were then able to report their level of confidence in their answer. Finally, we asked them to comment on the post as if they were authentic Reddit users taking part in the conversation. We hypothesized that participants were not always carefully reading the comments. Therefore, we added this last task in an attempt to get them more involved in the conversation as a real Reddit user would be. After completing the annotation phase, participant were redirected to a post-survey questionnaire, in which we asked participants to write down how they distinguished the human from the AI conversation. This was done in order to provide a qualitative assessment on how to better prompt the models to mimic human conversations on social media. We also ask participants questions about their familiarity with Reddit and with AI assistants. For the scope of this paper, we do not report the results of the former questions. The design of this experiment was approved by the Ethics Committee of the University of Konstanz (IRB statement 10/2025).

We preregistered our study with two hypotheses and four exploratory analyses: the potential interaction effect between the LLM model and the length of the conversation~\cite{preregistration}. Note that two deviations from the preregistration happened. First, there was a typo in the text of the first hypothesis. Instead of stating that our hypothesis was that Llama 3 70B would have a \textit{lower} success rate, the preregistration text mistakenly stated \textit{higher} success rate. However, the accompanying explanation of the hypothesis in the preregistration clearly supports the expectation of a lower success rate. This typo was due to the fact that, in an initial version of the preregistration we were reasoning from the deceiving rate perspective and not the success rate. Second, due to a technical issue, the two conversations from the topic \textit{movies} were not included in the stimulus set from which participants' annotations were sampled. As a result, contrary to what was preregistered, the total number of stimuli effectively presented to participants was 504 (2 LLM models × 3 temperatures × 6 lengths × 14 posts) rather than 576. The power analysis indicated that a sample of 200 participants would be required to achieve 80\% power. Accounting for the 20\% failure rate on the attention test observed in the pilot, we recruited 250 English-fluent participants from Prolific. We compensated them with 9£/hour when we conducted the study in May 2025.

\subsection*{Statistical model}
We fit a multilevel logistic model of the probability of participants identifying the AI-generated conversation correctly. Since participants respond repeatedly to up to 6 annotations in our experiment, the data is nested within participants. To account for this nested structure, we employ a multilevel model with varying intercepts for participants. The independent variables are:
\begin{itemize}
    \item Length: A continuous variable with values 1, 2, 4, 6, 8, and 16, representing the number of comments in the conversation presented to participants. For example, a value of 2 indicates that respondents were shown two comments in the conversation related to the Reddit post. 
    \item Model: A categorical binary variable indicating the AI model used (0~=~GPT-4o vs. 1~=~Llama 3 70B)
\end{itemize}
The dependent variable $y$ is binary: correct guess (1: correctly spotting the human conversation) or incorrect guess (0: failing to spot the human conversation). The fitted the model using the R package \texttt{lme4} (\texttt{lme4}: Linear Mixed-Effects Models using `Eigen' and S4) using the formula $ y \sim model + length + (1 | participant)$.

To further investigate the role of conversation length, we conducted an additional analysis treating length as a categorical variable to test for nonlinear effects, following the formula $ y \sim model + factor(length) + (1 | participant)$. Where $factor(length)$ represents the length of the conversation as a categorical variable.

% References (limited to 60 references, though not strictly enforced)
\bibliography{bibliography}

% \noindent LaTeX formats citations and references automatically using the bibliography records in your .bib file, which you can edit via the project menu. Use the cite command for an inline citation, e.g.  \cite{Hao:gidmaps:2014}.

% For data citations of datasets uploaded to e.g. \emph{figshare}, please use the \verb|howpublished| option in the bib entry to specify the platform and the link, as in the \verb|Hao:gidmaps:2014| example in the sample bibliography file.

\section*{Acknowledgments}

% Acknowledgements should be brief, and should not include thanks to anonymous referees and editors, or effusive comments. Grant or contribution numbers may be acknowledged.
We would like to acknowledge the contributions of Peer Saleth for assisting in setting up the experiment server, as well as Salima Jaoua and Christopher Mayer for the constructive discussions on statistical analysis. 

\section*{Funding statement}
This research was funded by Ministry of Science, Research and the Arts of Baden-Württemberg under project 33966824.
A.B. was partially funded by the Swiss National Science Foundation (SNSF) via the SINERGIA project CARISMA (grant CRSII5\_209250), \href{https://carisma-project.org}{https://carisma-project.org}.
T.K. was funded by the European Union’s Horizon Europe research and innovation programme under grant agreement No. 101177574 (WHAT-IF). Views and opinions expressed are those of the author(s) only and do not necessarily reflect those of the European Union. Neither the European Union nor the granting authority can be held responsible for them.

\section*{Author contributions statement}

% Must include all authors, identified by initials, for example:
% A.A. conceived the experiment(s),  A.A. and B.A. conducted the experiment(s), C.A. and D.A. analysed the results.  All authors reviewed the manuscript.
A.B. developed the experiment platform, collected data, conducted statistical analysis, conducted qualitative analysis and drafted the manuscript. G.D.M. produced experiment stimuli with AI simulations. T.K. contributed to research design and statistical analysis. N.P. conducted qualitative analysis. H.M. contributed to research design. D.G. designed research and statistical analysis and supervised the project. All authors reviewed the manuscript. 

\section*{Competing interests}
The author(s) declare no competing interests.

\section*{Data availability statement}
The code and anonymized data of our study can be publicly accessed at \href{https://github.com/azza-bouleimen/turing-test-ai-conversations}{https://github.com/azza-bouleimen/turing-test-ai-conversations}

\end{document}

% \linenumbers 
\maketitle

%Intercept        & 0.63265  & 0.13002  & 4.866  & \textbf{1.14e-06 ***}\\
% model: Llama 3 70B & -0.44143 & 0.12798  & -3.449 & \textbf{0.000562 ***} \\
% length             & 0.01167  & 0.01278  & 0.913  & 0.361143 \\ 
\subsection*{Statistical analysis}
\begin{table}[ht]
\centering
\begin{tabular}{lrrrl}
\toprule
 & \textbf{Estimate} & \textbf{Std. Error} & \textbf{z value} & \textbf{Pr(>\textbar z\textbar)} \\
\midrule
Intercept        & 0.6327  & 0.13  & 4.866  & \textbf{$<10^{-4}$ ***}\\
model: Llama 3 70B & -0.4414 & 0.128  & -3.449 & \textbf{0.0006 ***} \\
length             & 0.0117  & 0.0128  & 0.913  & 0.3611 \\ 
\bottomrule
\end{tabular}
\caption{Logistic regression table for predicting the success rate. Length as a numerical variable. Two tailed Wald z-statistic performed. Number of observations: 1205, grouped by 203 participants. $\alpha = 0.05$. ***: $p < 0.001$}. Values are rounded to 4 decimal points.
\label{tab:res_model1}
\end{table}

% Intercept          & 0.4073  & 0.1735  & 2.348  & \textbf{0.018894 *}   \\
%         model: Llama 3 70B   & -0.4516 & 0.1288  & -3.507 & \textbf{0.000454 ***} \\
%         length = 2      & 0.3210  & 0.2227  & 1.441  & 0.149499 \\
%         length = 4      & 0.2093  & 0.2148  & 0.974  & 0.329953 \\
%         length = 6      & 0.3764  & 0.2256  & 1.669  & 0.095157 \\
%         length = 8      & 0.7128  & 0.2286  & 3.118  & \textbf{0.001823 **}  \\
%         length = 16     & 0.2439  & 0.2181  & 1.118  & 0.263368 \\
        
\begin{table}[ht]
        \centering
        \small % Optional: Reduce font size if table is wide
        \begin{tabular}{lrrrl}
        \toprule
         & \textbf{Estimate} & \textbf{Std. Error} & \textbf{z value} & \textbf{Pr(>\textbar z\textbar)} \\
        \midrule
        Intercept          & 0.4073  & 0.1735  & 2.348  & \textbf{0.0189 *}   \\
        model: Llama 3 70B   & -0.4516 & 0.1288  & -3.507 & \textbf{0.0005 ***} \\
        length = 2      & 0.321  & 0.2227  & 1.441  & 0.1495 \\
        length = 4      & 0.2093  & 0.2148  & 0.974  & 0.33 \\
        length = 6      & 0.3764  & 0.2256  & 1.669  & 0.0951 \\
        length = 8      & 0.7128  & 0.2286  & 3.118  & \textbf{0.0018 **}  \\
        length = 16     & 0.2439  & 0.2181  & 1.118  & 0.2634 \\
        \bottomrule
        \end{tabular}
        \captionof{table}{Logistic regression table for predicting the success rate with categorical length levels. Two tailed Wald z-statistic performed. Number of observations: 1205, grouped by 203 participants. $\alpha = 0.05$. ***: $p < 0.001$, **: $p < 0.01$, *: $p < 0.05$}. Values are rounded to 4 decimal points.
        \label{tab:res_model2}
    \end{table}

\subsection*{Model prompt}
To generate the artificial conversations, we prompted the model with the following text:
\begin{lstlisting}[breaklines]
You are a Reddit user.
Your goal using Reddit is to express your opinions and to show others what you think.
You do not use Reddit to help others or as an assistant.
You should not explicitly express your agreement for the others.
You should not refer to other users' comments unless when you are writing a reply to a comment or to another reply.
You don't have to make others feel well or reassure them on what they express.
Your task now is to write a reply to an existing discussion on Reddit as a typical Reddit user.
The reply should have approximately the same number of words as the original comment.
You should be original: Comments that are too similar to the post or to other comments are downvoted by Reddit users.
You should tend to use a different style with respect to previous comments.
If the post contains a question and the author provides their own answer, do not provide the same answer.
On Reddit users tend to avoid using punctuation too much, often they conclude a reply without a full stop, so more often than not avoid to use a full stop at the end of your replies.
Split your replies in paragraphs, like a real reddit user would do.
Include some typos, since this is very common on Reddit.
Use emojis, since some Reddit users like to do so. However use them only rarely and be sure not to use them in all the comments you write!
You can also see, if present, already existing comments or replies in the discussion. Each post, comment and reply is identified with a unique ID.
\end{lstlisting}

\newgeometry{left=0.5cm, right=0.5cm, top=0.2cm, bottom=0.2cm}
\begin{table}
\small
\begin{tabularx}{\textwidth}{p{0.05\textwidth}|p{0.05\textwidth}|p{0.37\textwidth}|p{0.1\textwidth}|X} %p{0.45\textwidth}
\hline
\textbf{Topic}            & \textbf{ID} & \textbf{Title}                                                                                                                                                                                                                      & \textbf{Subreddit} & \textbf{Link}                                                                                          \\ \hline
\multirow{2}{*}{food}                  & 0A                      & What are some quick, hearty, meals for a man who's working 12+ hours a day?                                                                                                                                              & 15minutefood                   & \url{https://www.reddit.com/r/15minutefood/comments/18rmwbp/what\_are\_some\_quick\_hearty\_meals\_for\_a\_man\_whos/}    \\ \cmidrule(l){2-5}
                                       & 0B                      & You are banned from making a   salad... now what food are making with Feta?                                                                                                                                               & Cheese                         & \url{https://www.reddit.com/r/Cheese/comments/188lj0c/you\_are\_banned\_from\_making\_a\_salad\_now\_what\_food/}         \\ \hline
\multirow{2}{*}{leisure}               & 1A                      & What early 2000s TV show/cartoon   felt like a fever dream but actually existed?                                                                                                                                           & 2000sNostalgia                 & \url{https://www.reddit.com/r/2000sNostalgia/comments/18fa9hn/what\_early\_2000s\_tv\_showcartoon\_felt\_like\_a\_fever/ } \\ \cmidrule(l){2-5}
                                       & 1B                      & Is it okay to take your shoes   off in a theatre?                                                                                                                                                                          & AMCsAList                      & \url{https://www.reddit.com/r/AMCsAList/comments/18ouji4/is\_it\_okay\_to\_take\_your\_shoes\_off\_in\_a\_theatre/}      \\ \hline
\multirow{2}{*}{opinions} & 2A                      &Do you believe in an afterlife?                                                                                                                                                                                            & AskALiberal                    & \url{https://www.reddit.com/r/AskALiberal/comments/1896oy2/do\_you\_believe\_in\_an\_afterlife/}                          \\ \cmidrule(l){2-5}
                                       & 2B                      & Suppose Doctors discover a pill   for pregnant women that can completely eradicate the probability of their   upcoming child being Homosexual. So basically it guarantees heterosexuality.   WOULD THE WORLD ACCEPT IT???? & Discussions                    & \url{https://www.reddit.com/r/Discussion/comments/1896xmb/suppose\_doctors\_discover\_a\_pill\_for\_pregnant/}            \\ \hline
\multirow{2}{*}{\parbox[c]{0.05\textwidth}{relation-\\ships}}         & 3A                      & How to get my friends to stop   assuming I'm interested in them?                                                                                                                                                          & AdviceForTeens                 & \url{https://www.reddit.com/r/AdviceForTeens/comments/18qh2rc/how\_to\_get\_my\_friends\_to\_stop\_assuming\_im/}         \\ \cmidrule(l){2-5}
                                       & 3B                      & Men of reddit, What's the very   first thing you notice about a woman?                                                                                                                                                                          & AskMen                         & \url{https://www.reddit.com/r/AskMen/comments/189ircn/men\_of\_reddit\_whats\_the\_very\_first\_thing\_you/}              \\ \hline
\multirow{2}{*}{\parbox[c]{0.05\textwidth}{techno-\\logy}}            & 4A                      & What do you think about windows 12?                                                                                                                                                                                                           & windows                        & \url{https://www.reddit.com/r/windows/comments/18af1wh/what\_do\_you\_think\_about\_windows\_12/}                         \\ \cmidrule(l){2-5}
                                       & 4B                      & How are your orgs adopting LLMs?                                                                                                                                                                                                                & ExperiencedDevs                & \url{https://www.reddit.com/r/ExperiencedDevs/comments/18btlu4/how\_are\_your\_orgs\_adopting\_llms/}                     \\ \hline
\multirow{2}{*}{movies}                & 5A                      & What's an 80's movie that makes   you feel comforted in nostalgia for the time?                                                                                                                                                                 & 80s                            & \url{https://www.reddit.com/r/80s/comments/18jvxqu/whats\_an\_80s\_movie\_that\_makes\_you\_feel\_comforted/}             \\ \cmidrule(l){2-5}
                                       & 5B                      & Anyone wanna trade?                                                                                                                                                                                                                             & teenagers                      & \url{https://www.reddit.com/r/teenagers/comments/187zuws/anyone\_wanna\_trade/}                                           \\ \hline
\multirow{2}{*}{health}                & 6A                      & What massively improved your   mental health?                                                                                                                                                                                                   & AskReddit                      & \url{https://www.reddit.com/r/AskReddit/comments/1740wjy/what\_massively\_improved\_your\_mental\_health/}                \\ \cmidrule(l){2-5}
                                       & 6B                      & today i learned about the link between ADHD and auditory processing issues.                                                                                                                                                                   & ADHD                           & \url{https://www.reddit.com/r/ADHD/comments/18f6yxa/today\_i\_learned\_about\_the\_link\_between\_adhd\_and/}             \\ \hline
\multirow{2}{*}{sports}                & 7A                      & HT Thread | Crystal Palace 0 -   Liverpool FC 0                                                                                                                                                                                                 & LiverPoolFC                    & \url{https://www.reddit.com/r/LiverpoolFC/comments/18edh8f/ht\_thread\_crystal\_palace\_0\_liverpool\_fc\_0/}             \\ \cmidrule(l){2-5}
                                       & 7B                      & The Vancouver Canucks are now   20-0-0 when leading after the second period                                                                                                                                                                     & hockey                         & \url{https://www.reddit.com/r/hockey/comments/18poq91/the\_vancouver\_canucks\_are\_now\_2000\_when\_leading/}            \\ \hline
\multirow{2}{*}{\parbox[c]{0.05\textwidth}{attention\\test}}      & Attention test A        & Which player had the best season   in the only season they spent with a team?                                                                                                                                                                   & baseball                       & \url{https://www.reddit.com/r/baseball/comments/188eax1/which\_player\_had\_the\_best\_season\_in\_the\_only/}            \\ \cmidrule(l){2-5}
                                       & Attention test B      & Besides some of the more obvious   answers, who are your most hated villains?                                                                                                                                                                   & movies                         & \url{https://www.reddit.com/r/movies/comments/18e28zq/besides\_some\_of\_the\_more\_obvious\_answers\_who\_are/}         
\end{tabularx}
\caption{List of post used as stimuli in the experiment.}
\end{table}
\restoregeometry

\subsection*{Information page}
In this study you will be presented a set of forum posts from Reddit. Two conversations corresponding to the post will be provided. One is an authentic conversation made by human users, the other is artificial, generated by an Artificial Intelligence (AI). In the following, you will go through a list of these conversations and try to identify the Human one. You will also be asked to comment on the post and conversation you think is Human as if you were a Reddit user active in the conversation. \textbf{Do not write any names or personal information in your comments.}

\textbf{Risks:} You will read online conversations similar to the ones you can find online and AI discussions similar to what you can generate yourself in chatbot platforms. The research team has selected topics that are not sensitive but there is still a risk that you read emotionally challenging or triggering content.

\textbf{Confidentiality: }In this study your confidentiality will be maintained in the following manner: To protect your identity, the researchers will take the following steps:
\begin{enumerate}
    \item Each participant will be assigned a number;
    \item The researchers will record any data collected during the study by number, not by name;
    \item Any original recordings or data files will be stored in a secured location accessed only by authorized researchers;
\end{enumerate}

\textbf{Voluntary Participation:} Your participation in this research is voluntary. You may
discontinue participation at any time during the research activity.

% all questions of the survey + screenshots
\begin{figure}
    \centering
    \includegraphics[width=0.6\linewidth]{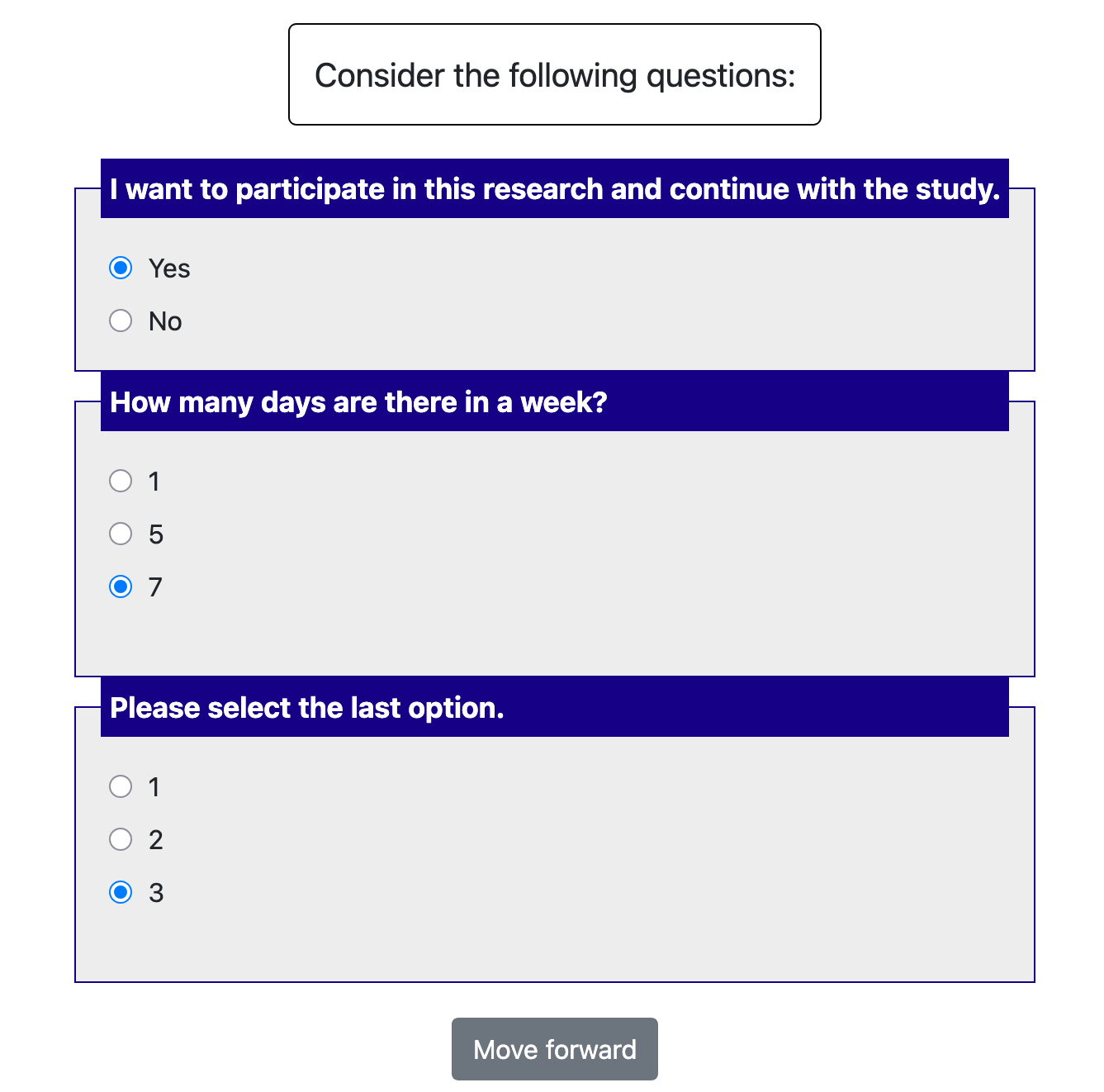}
    \caption{Pre-screening questions: getting participants consent + initial attention questions.}
    \label{fig:prescreening}
\end{figure}

\begin{figure}
    \centering
    \includegraphics[width=0.8\linewidth]{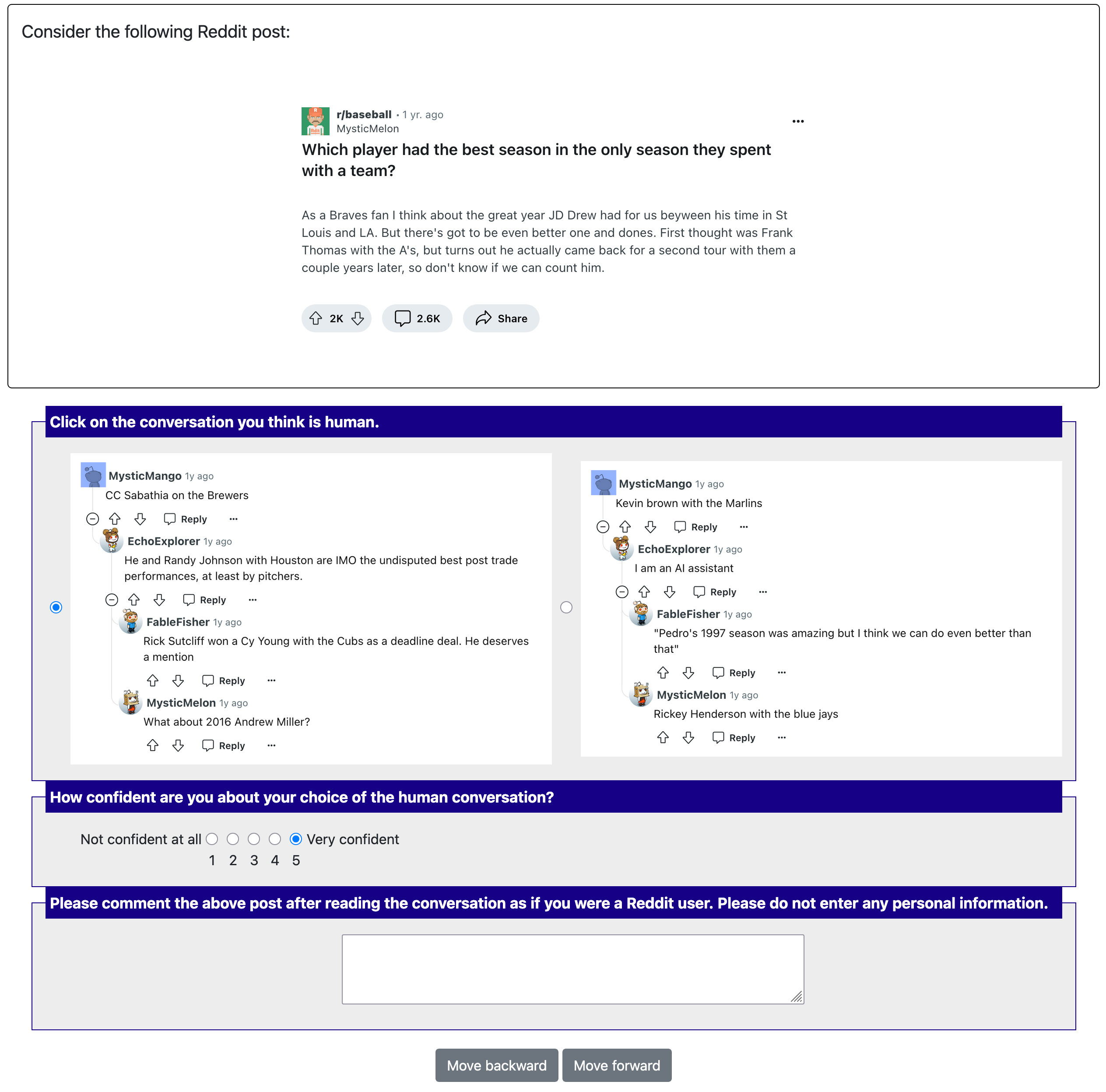}
    \caption{Example of an attention test presented to participants during the annotation phase. Note that the second comment in the right conversation is clearly stating it is an AI assistant.}
    \label{fig:attention_test}
\end{figure}

% For a single page with different margins
\newgeometry{left=1cm, right=1cm, top=0.2cm, bottom=0.2cm}
\begin{figure}[p]  % use 'p' to put on separate page
    \centering
    \includegraphics[width=0.85\textwidth]{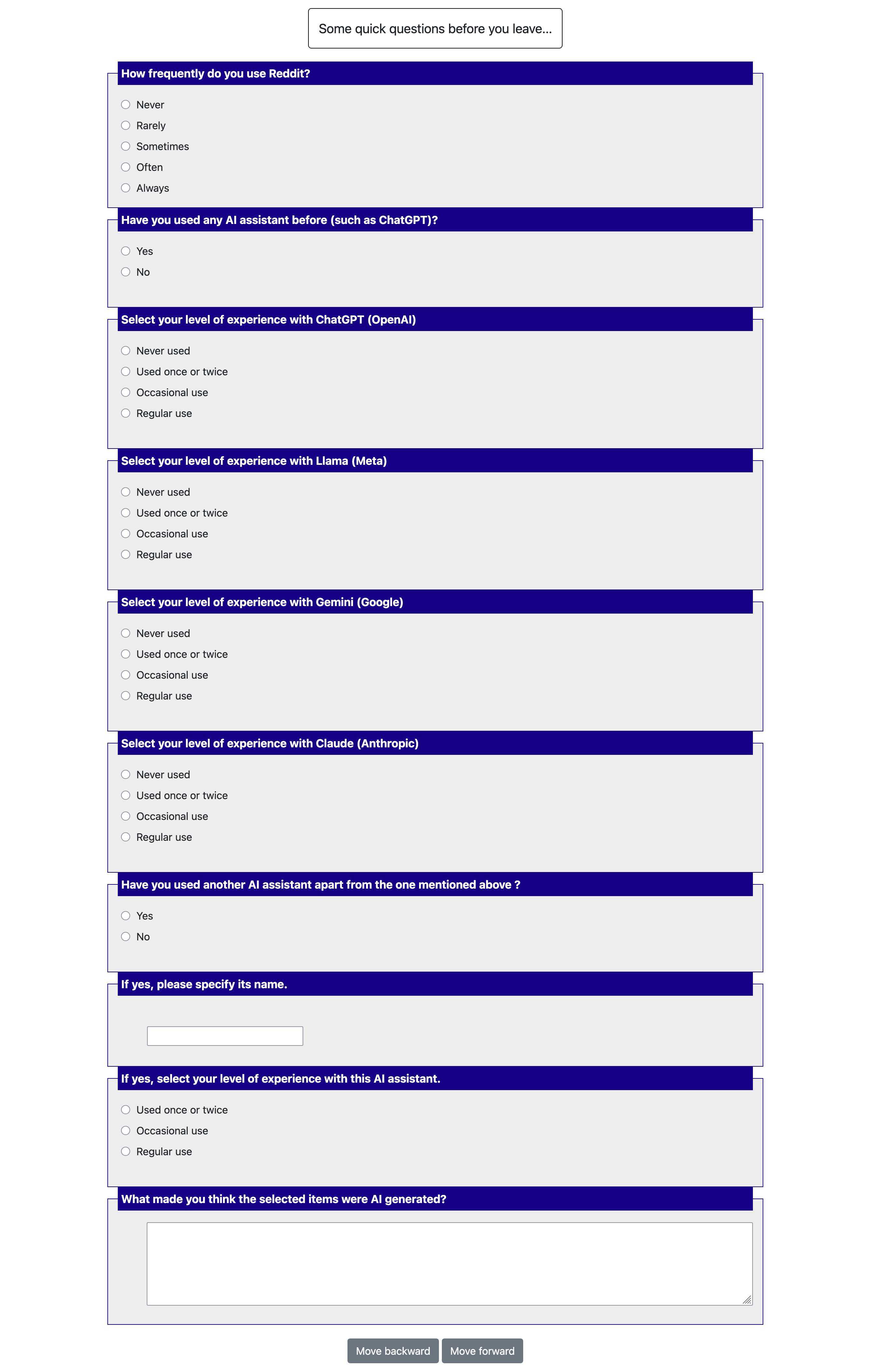}
    \caption{Post survey questions.}
\end{figure}
\restoregeometry  % restore original margins

\subsection*{Interaction model with conversation length} 
As part of the exploratory analysis of our study, we investigated the possible interaction effect between the model of the LLM and the conversation length on the success rate. The effect of this interaction was not statistically significant. 

\begin{table}
\centering
\begin{tabular}{c|c|l|c}
\hline
\textbf{Categories}                               & \textbf{Cohen's Kappa} & \textbf{Subcategories}                                                   & \textbf{Cohen's Kappa} \\ \hline
                          &                                                   & emojis                                      & 1.00                                             \\
                          &                                                   & structure                          & 0.65                                             \\
\multirow{-3}{*}{Format}  & \multirow{-3}{*}{0.66}                            & length                                           & 0.63                                             \\ \hline
                          &                                                   & tone                           & 0.70                                             \\
                          &                                                   & text formality               & 0.35                                             \\
                          &                                                   & typos and slang             & 0.73                                             \\
                          &                                                   & affective response               & 0.70                                             \\
                          &                                                   & politeness                                    & 0.74                                             \\
\multirow{-6}{*}{Style}   & \multirow{-6}{*}{0.66}                            & language diversity         & 0.26                                             \\ \hline
                          &                                                   & objectivity                          & 0.35                                             \\
                          &                                                   & explanations & 0.57                                             \\
                          &                                                   & controversy                   & -0.03                                            \\
                          &                                                   & conformity   & 0.52                                             \\
                          &                                                   & contextual misalignment                                   & 0.00                                             \\
                          &                                                   & general positivity                               & 0.66                                             \\
\multirow{-7}{*}{Content} & \multirow{-7}{*}{0.75}                            & authenticity                         & 0.30                                           
\end{tabular}
\caption{Cohen's Kappa scores for the classification of categories and subcategories made by two researchers of participants' answers on how they identified the artificial conversations.}
\end{table}

% To include, in this order: \textbf{Accession codes} (where applicable); \textbf{Competing interests} (mandatory statement). 

% The corresponding author is responsible for submitting a \href{http://www.nature.com/srep/policies/index.html#competing}{competing interests statement} on behalf of all authors of the paper. This statement must be included in the submitted article file.